\documentclass[sigconf]{acmart}
\usepackage{array}
\usepackage{pifont}
\usepackage{hyperref}
\usepackage{tcolorbox}
\usepackage{color}
\usepackage[subtle]{savetrees}

\copyrightyear{2024}
\acmYear{2024}
\setcopyright{acmlicensed}
\acmConference[CSCS '24]{Proceedings of the 2024 Cyber Security in CarS Workshop }{October 14--18, 2024}{Salt Lake City, UT, USA}
\acmBooktitle{Proceedings of the 2024 Cyber Security in CarS Workshop (CSCS '24), October 14--18, 2024, Salt Lake City, UT, USA}
\acmDOI{10.1145/3689936.3694694}
\acmISBN{979-8-4007-1232-6/24/10}



\begin{abstract}
In the current landscape of autonomous vehicle (AV) safety
and security research, there are multiple isolated
problems being tackled by the community at large.
Due to the lack of common evaluation criteria, several
important research questions are at odds with one another.
For instance, while much research has been conducted on
physical attacks deceiving AV perception systems, there is
often inadequate investigations on working defenses and
on the downstream effects of safe vehicle control.

This paper provides a thorough description of the current
state of AV safety and security research.
We provide individual sections for the primary research questions
that concern this research area, including AV surveillance,
sensor system reliability, security of the AV stack, algorithmic
robustness, and safe environment interaction.
We wrap up the paper with a discussion of the
issues that concern the interactions of these separate problems.
At the conclusion of each section, we propose future research
questions that still lack conclusive answers. This position
article will serve as an entry point to novice and veteran
researchers seeking to partake in this research domain.
\end{abstract}

\begin{document}

\title{Achieving the Safety and Security of the End-to-End AV Pipeline}

\author[N. T. Curran]{Noah T. Curran}
\orcid{0009-0007-8339-9297}
\email{ntcurran@umich.edu}
\affiliation{%
  \institution{University of Michigan}
  \city{Ann Arbor}
  \state{MI}
  \country{USA}
  }

\author[M. Cho]{Minkyoung Cho}
\orcid{0009-0003-6979-0303}
\email{minkycho@umich.edu}
\affiliation{%
  \institution{University of Michigan}
  \city{Ann Arbor}
  \state{MI}
  \country{USA}}

\author[R. Feng]{Ryan Feng}
\orcid{0000-0002-4767-274X}
\email{rtfeng@umich.edu}
\affiliation{%
  \institution{University of Michigan}
  \city{Ann Arbor}
  \state{MI}
  \country{USA}}

\author[L. Liu]{Liangkai Liu}
\orcid{0000-0002-6149-9859}
\email{liangkai@umich.edu}
\affiliation{%
  \institution{University of Michigan}
  \city{Ann Arbor}
  \state{MI}
  \country{USA}}

\author[B. J. Tang]{Brian Jay Tang}
\orcid{0000-0001-6140-922X}
\email{bjaytang@umich.edu}
\affiliation{%
  \institution{University of Michigan}
  \city{Ann Arbor}
  \state{MI}
  \country{USA}}

\author[P. MohajerAnsari]{Pedram MohajerAnsari}
\orcid{0009-0001-6033-6759}
\email{pmohaje@clemson.edu}
\affiliation{%
  \institution{Clemson University}
  \city{Clemson}
  \state{SC}
  \country{USA}}

\author[A. Domeke]{Alkim Domeke}
\orcid{0009-0005-2714-2011}
\email{adomeke@clemson.edu}
\affiliation{%
  \institution{Clemson University}
  \city{Clemson}
  \state{SC}
  \country{USA}}

\author[M. D. Pesé]{Mert D. Pesé}
\orcid{0000-0001-9192-5823}
\email{mpese@clemson.edu}
\affiliation{%
  \institution{Clemson University}
  \city{Clemson}
  \state{SC}
  \country{USA}}

\author[K. G. Shin]{Kang G. Shin}
\orcid{0000-0003-0086-8777}
\email{kgshin@umich.edu}
\affiliation{%
  \institution{University of Michigan}
  \city{Ann Arbor}
  \state{MI}
  \country{USA}}
  
\renewcommand{\shortauthors}{Noah T. Curran et al.}

\maketitle






\section{Introduction}
\label{sec:intro}
As part of the proliferation of autonomous vehicles (AVs),
automotive manufacturers have argued that they will
improve the overall safety of 
roadways~\cite{cruise2023safety,waymo2023safety}.
For instance, after one-million driverless miles,
Cruise has claimed that their ``safety-
minded"~\cite{cruise2023safetyreport}
self-driving technology is safer on average than a
human driver~\cite{cruise2023onemillion}.
Despite this, Cruise has faced immense scrutiny over
their self-driving technology due to a series of recent
accidents~\cite{hawkins2023robotaxis,shakir2023robotaxis,mickle2023gm}.
Some question whether a human driver would have
caused these accidents~\cite{koopman2023safety}.
In the aftermath, Cruise has recalled their AVs and
ceased operation for the foreseeable
future~\cite{koopman2023standdown,cruise2023important}.

Alongside the safety issues that AVs must still overcome,
there have been advancements from academia of theoretical
security concerns of AVs. These findings range from attacks
on perception systems using adversarial ML techniques to
tricking the object tracking systems. There are also more
traditional attack vectors that use software exploits to gain
unauthorized privilege escalation within the AV software
stack. Coupling both the security and safety of an AV,
assessing the robustness of the guarantees of correct
AV operation is a complex procedure.

\begin{figure}[t]
    \centering
    \includegraphics[width=0.7\columnwidth]{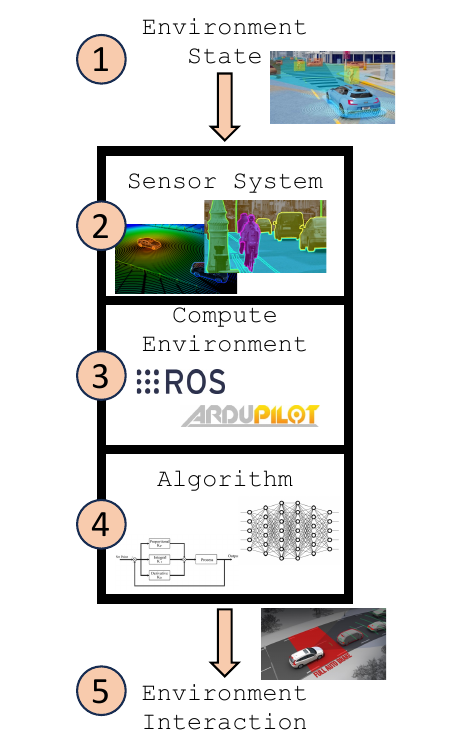}
    \caption{A summary of the components of the AV system.
    The focus is on the major topics of inquiry that concern
    the safety/security community. (1) Preserving the privacy
    of the environment around the vehicle. (2) Ensuring the
    correctness of sensor measurements. (3) Maintaining the
    integrity of the compute environment stack, including the
    middleware, RTOS, and CPU/GPU resources. (4) Proving
    algorithmic guarantees/robustness with respect to ensuring
    AV mission completion. (5) Providing safe and socially acceptable
    control of the AV as it interacts with its environment.}
    \label{fig:overview}
    \Description[5 stage pipeline]{The 5 stages are in a column, ordered from top to bottom. It orders 1 environment state, 2 sensor system, 3 compute environment, 4 algorithm, 5 environment interaction.}
    \vspace{-0.2in}
\end{figure}

The structure of our position paper is as follows.
First, we provide a brief background of concepts
necessary for understanding AV safety and security.
Then, we provide the current state of research from
each identifiable community in the literature (summarized
in \autoref{fig:overview}). Each of these sections concludes
with the limitations and takeaways.

In this article, we contribute the following:
\begin{itemize}
    \item We provide a high-level overview of each
        major component to the AV. These components
        are summarized in \autoref{fig:overview}.
    \item For each component of the AV,
        we present limitations of the current
        state-of-the-art research which has
        direct implications toward safety/security.
    \item To jump-start future research that
        may prioritize these limitations, we propose
        research questions for the major components
        of the AV, as well as the end-to-end system
        as a whole.
\end{itemize}

\section{Background}
\label{sec:background}
In \autoref{fig:overview} we present a
non-traditional overview of the components of
the AV. Typically, most introductions to AVs
present a \textit{Sense-Plan-Act} control loop.
However, \textit{Sense-Plan-Act} does not have an
abstraction that shows the important portions
of the attack surface. In contrast, our
representation combines the planning and control
of the AV as part of the algorithmic processes,
and we expose the middleware as a component between
the sensor system and these algorithms.
Our abstraction also raises importance to the
environment before and after the AV is controlled.
In this section we provide background necessary for
understanding the goals of these components before
we discuss the safety and security issues in the
following sections.

\subsection{Perception Sensors}
In order to know what the world looks like around
itself, the AV must use sensors that provide perception.
The primary sensing units of the vehicle may include
one or more camera, radar, LiDAR,
or ultrasonic sensor~\cite{foresight2023overview}.
For various other control or infotainment functions,
the vehicle may also have a GNSS, gyroscope, accelerometer,
microphone~\cite{ignatious2022overview}, or wireless
transmitter/reciever~\cite{curran2023wip}.
On the same vehicle, there may be
different forms of these sensors, such as thermal,
monoscopic, or stereoscopic cameras; there are short-range
and long-range radars; and LiDARs can use laser pulses,
a continuous laser beam, a rotating laser beam, or solid
state. The arrangement of the perception sensors
will depend on the needs and requirements of the AV
designer. The perception system may include a component
for processing the data and extracting information critical
to the operation of the vehicle. For instance, it may parse
image data for important objects or segment it entirely into
different critical regions. This data is then passed on
to the safety-critical tasks that need it.

\subsection{Middleware}
Communication middleware is a foundational layer that facilitates
efficient and reliable communication between the
various software and hardware components of an AV.
For instance, the various sensors are connected to
the middleware, which manages and directs the sensor
information toward the various tasks that require it.
The middleware will also provide priority assignments
to various tasks, as well as priviledge management
for accessing information and overridding task assignment.
The middleware also provides basic functionality for
scheduling the tasks and meeting real-time requirements.
These tasks may include algorithms that support safety or
comfort features that may autonomously operate the vehicle.
A few examples of a middleware include the
Robot Operating System (ROS)~\cite{quigley2009ros,macenski2022robot},
Ardupilot~\cite{ardupilot}, PX4~\cite{px4},
Yet Another Robot Platform (YARP)~\cite{yarp},
Cyber RT~\cite{cyber-rt}
and OpenRTM-aist~\cite{openrtmaist}.

\subsection{AV Tasks}
In the context of an AV, a task could be as simple as
checking the distance of the vehicle to nearby objects,
or as complex as a multi-object tracking algorithm.
These tasks each have their expected execution time,
deadline, and release time. Using this information,
each task is assigned a priority so that the middleware
knows which to execute next. Tasks that have safety-critical
deadlines, those that have catastrophic consequences if missed,
are called hard deadlines. Some deadlines may be missed
without consequence, and are called soft deadlines.
Those deadlines that do not lead to catastrophe if missed
but cause the result of the task to be unusable are called
firm deadlines. Depending on the task and environmental conditions,
these deadline concepts may have some applicable variation,
such as an $(m,k)$-firm deadline where at least $m$ out of
$k$ instances of a task must complete execution
before the deadline~\cite{hamdaoui1995dynamic}.

While the middleware can make
assumptions about the tasks in order to quickly obtain a
feasible schedule, they often have complex relationships
with one another. For instance, some tasks may have higher
priority based on the context (mixed-criticality system)~\cite{de2009scheduling},
may not be executable until another dependent task
completes~\cite{david2001jitter}, or may access a
resource shared by another task~\cite{easwaran2009resource}.
AVs encounter variable and complex environments,
so based on the context the middleware may be required
to enforce strict deadlines on specific tasks.
Thus, flexibility of task priority assignment is necessary
in order to maintain safe operation of the vehicle.

\section{Environment State}
\label{sec:pipeline_1}

For AVs, the environment state is crucial for vehicles' operation and interaction with their surroundings. The environment state refers to the comprehensive and real-time representation of all relevant factors and conditions surrounding the vehicle. This includes static elements such as road layouts, traffic signs, barriers, and driving setting as well as more dynamic elements such as current vehicle position, other vehicles, pedestrians, and weather conditions. An AV must accurately perceive and interpret the environment state as every action the AV takes, from changing lanes to adjusting speed, relies on an accurate representation of its current environment. For instance, the AV must be able to recognize a pedestrian crossing the street, understand that it signifies a need to slow down or stop, and take appropriate action.

\subsection{Driving Environments}

AVs drive through a vast number of public and private spaces, requiring them to navigate serious safety, security, and privacy concerns. At near-zero error rates, a level 4/5 AV is expected to handle complex contexts like urban landscapes and harsh weather. This leads to a large operational design domain, and AV companies utilize a find-and-fix approach when handling novel crash scenarios. The preference of this approach is due to trends in deep-learning approaches to use big data to improve the performance of models; the developers of the AVs are incentivized to maximize their data collection activities to improve training and evaluation sets with respect to minimizing accident statistics. To this end, AVs are equipped with information-rich and diverse sensors, making them capable of collecting incredibly sensitive data, particularly if driving scenes from multiple vehicles are collated.

\subsection{Surveillance}

Such sensor networks can be misused to perform targeted and/or mass surveillance~\cite{glancy2012privacy} of individuals or communities. Already, car companies collect and sell copious amounts of sensitive and personal data with 3rd parties with little to no data controls or opt-outs for users~\cite{mozilla2023carprivacy}. In some instances, law enforcement and government organizations have requested self-driving car data without warrants~\cite{Love2023Jun}. This may be especially concerning if AV data is collected and aggregated from multiple vehicles. AV systems may become an omnipresent platform for collecting and inferring people's activities, a notion that many people expressed overwhelming discomfort about in a study by Bloom \textit{et al.}~\cite{bloom2017self}. Continuously monitoring people for the purposes of law enforcement or advertising seriously threatens personal autonomy and creates technological platforms ripe for the abuse of power~\cite{leon2019eyes}. 

\subsection{Limitations \& Takeaways}

\subsubsection{Existing Privacy Research}
In the domain of AVs, there has been a serious lack of works discussing potential privacy risks and protection methods. Most AV privacy literature present legal arguments and concerns. While there is a heavy emphasis with research on location privacy and biometric (face) privacy, little research has investigated privacy solutions in vehicles (e.g., opting out, privacy-aware data collection and sensing, etc.).

\subsubsection{Proprietary ML Attacks}
Xie \textit{et al.} propose categorizing privacy risks into individual (people, attributes), population (communities, towns, demographics), and proprietary (ML models, architectures, sensor details), creating a taxonomy of attacks and defenses~\cite{xie2022privacy}. They, like most other works, tend to focus on adversarial ML attacks, (e.g., membership inference), differential privacy, or federated learning.~\cite{jallepalli2021federated,hassan2019differential,ghane2020preserving,pokhrel2020decentralized}. 

\subsubsection{Sensor Privacy}
We argue that this focus on proprietary privacy in AVs is insufficient, impractical, and fails to address the problem at its root cause -- intrusive data collection and processing practices. Rather, privacy protection mechanisms should be integrated earlier in AV pipelines, e.g., sensor systems, to prevent the identification of individuals. Rather than raw camera data being collected and processed by AV companies, the data could be processed on-vehicle via semantic segmentation or using obfuscation techniques.

\subsubsection{Usable Privacy Solutions}
Moreover, effective and strict privacy regulations on data collection and storage are needed as AV systems mature and become deployed in our society. Without these protections, end-users could resort to using anti-surveillance technologies/wearables that attack object detectors or sensors, increasing safety risks~\cite{wu2020making,xu2020adversarial}.



\begin{tcolorbox}[width=\columnwidth, colback=black!10, arc=3mm]
    \underline{\textsc{Research Questions}$-$\ding{182}}:
        \textit{To what extent do AVs exacerbate surveillance? What are the direct implications of widespread AV adoption, and how will it impact personal freedoms? Can we provide privacy to the environment surrounding an AV without detriment to the AV's performance (safety and otherwise)?}
\end{tcolorbox}

\section{Sensor System}
\label{sec:pipeline_2}

The perception modules in AVs are responsible for
understanding and interpreting the environment.
Typical mechanisms of self-driving vehicles involve
perceiving surroundings, tracking objects, and planning
future trajectories for the vehicle itself. As a result,
the perception module is especially crucial as it provides
initial results for use in safety-critical algorithms,
which causes the perception module to be
a desirable target for an attacker. On one hand,
the adversary may alter the environment. The adversary
may also physically inject a signal directly into the
sensor. We categorize these as {\em Indirect} and
{\em Direct AV Perception Attacks} respectively.

\subsection{Indirect AV Perception Attacks}
For {\em indirect} perception attacks, there are
several proven methods from prior work. For instance,
an adversary may introduce modifications to the road
pavement to
cause the vehicle to misinterpret its lane position and
in turn mislead the lane detection
system~\cite{jing2021too,sato2021dirty}.
Another strategy employed is to project adversarial patches
to road signs and nearby
walls~\cite{lovisotto2021slap,man2023person,sato23infrared}.
The creation of adversarial objects has also shown
effectiveness in misguiding object detection
systems~\cite{cao2021invisible}. Additionally,
pedestrians may wear clothes with materials
or patterns specifically designed to trick
the sensors in order to remain
undetected~\cite{zhu2022infrared} or detected with incorrect
label~\cite{bandara2023this,hu2022adversarial,xu2020adversarial}.


\subsection{Direct AV Perception Attacks}
When an adversary performs a {\em direct}
perception attack, it involves physically
manipulating the sensor to provide
falsified measurements. These attacks
may be naive with the only goal being
to obfuscate the environment. However,
an advanced adversary is capable of
removing and/or adding
specific objects without impacting the rest of the
environment. These attacks have shown
to at least reduce the overall performance of
ML models that AVs use, or in the worst case
cause a collision of the vehicle.
One way of directly causing these attacks is
for an adversary to use a laser to inject
a desired signal into the sensor, such
as LiDAR removal~\cite{cao2021invisible,hau2021object,tu2020physically,zhu2021can,cao2023you}
and appearing~\cite{cao2021demo,cao2019adversarial,shin2017illusion,sun2020towards}
attacks. To a lesser degree, lasers have been
shown to be useful on other sensors as well,
such as microphones~\cite{sugawara2020light}.
While lasers have not been shown
to be effective for these purposes on
cameras~\cite{hallyburton2022security},
infrared light~\cite{sato2024invisible,sato23infrared,wang2021can},
acoustic signals~\cite{ji2021poltergeist},
and electromagnetic interference~\cite{kohler2022signal}
have been shown to be capable of directly injecting
adversarial signals into the camera. Ultrasonic
sensor spoofing has also been demonstrated to
be feasible~\cite{yan2016can}.





\subsection{Defenses}
To counter {\em indirect} perception attacks,
state-of-the-art defenses suggest training 
the lane detection system on a diverse dataset that includes
various perturbations, enabling the defensive module to
recognize and ignore deceptive patterns. Adversarial training,
a technique that has gained traction in recent years,
specifically addresses attacks that use
adversarial patches~\cite{lovisotto2021slap, gehr2018ai2, cohen2019certified}.
By training the vehicle's neural networks with
adversarial examples, the system becomes more robust
to deceptive
inputs~\cite{xu2020adversarial,zhu2021adversarial,sun2020towards}.
While this approach might
marginally decrease the system's performance under
normal conditions, its strength against known
attacks is significantly enhanced.

Defenses against {\em direct} perception attacks
are trickier due to a stronger and more resourceful adversary. 
These defenses may go as far as requiring hardware modifications
to thwart the attacker. For instance, defenses for LiDAR
removal attacks search for ``shadows" left behind as an
artifact of the attack~\cite{wang2021can,hau2022using}.
An adaptive adversary may attempt to hide the removal of an
object by covering the shadow with a spoofed signal.
There are also defenses for the appearing attacks, which
take advantages of constraints imposed on the adversary
for launching the attack~\cite{xiao2023exorcising,cho2023adopt}.

Filters are also used to defend against the {\em direct}
attacks, such as a software filter for detecting infrared
light attacks~\cite{wang2021can}. Hardware filters or shields
are suggested for defense against attacks on MEMS
accelerometers~\cite{trippel2017walnut}
and gyroscopes~\cite{guri2022gairoscope}. Physical filters
may be useful to reduce the physical surface area an
attack must target. For instance, camera hoods or diffracting
films can block straight light beams from
laser attacks~\cite{sugawara2020light}.

Finally, a common approach is to use traditional
intrusion detection systems to detect anomalous sensor data.
Such a system may use estimates of sensor data to compare with
the measurements~\cite{curran2023using,curran2023boeing} or
use a deep neural network to learn hidden features of anomalous
sensor data~\cite{zhang2021deep}.

\subsection{Limitations \& Takeaway}

\subsubsection{Absence of Provable Defenses}
Often defenses against attacks on the perception
system are constantly in a back-and-forth game of
cat-and-mouse. Each time a new attack is defended
against, a new attack takes its place to circumvent
the defense. While defenses often remain under-evaluated,
presented as an afterthought to the attack
component, the primary issue is that defenses are
typically hard to prove as secure. Oftentimes, defenses
use probabilistic approaches that are still vulnerable
to clever deception.

\subsubsection{No Common Evaluation Practice}
Despite the growing popularity of sensor attacks, the
metrics used to evaluate success vary substantially
across works. In some cases, simply achieving a
misclassification is enough, while in other cases
evading detection from the most popular defense
is sufficient. While the difference in goal is
largely due to the ``story" a work builds, this
makes comparing the impact of novel attacks
impossible. This leads to our next takeaway.

\subsubsection{Rarely Known Downstream Effect on Safety}
The lack of common evaluation practice has caused
a departure from focusing on actual AV safety as a
consequence of a novel attack. While the attacks
thoroughly evaluate the initial goal, it is not
understood to what effect the attacks cause the AV
to actually do anything dangerous. In a few work,
dangerous control is shown to be feasible, but there
is a notable absence of the range of scenarios
for which this dangerous control may occur. Future
work can improve the state of the art by providing
a rigorous analysis of the state-space for which
sensor attacks may successfully cause dangerous 
control outcomes.

\begin{tcolorbox}[width=\columnwidth, colback=black!10, arc=3mm]
    \underline{\textsc{Research Question}$-$\ding{183}}:
        \textit{What set of evaluation metrics may satisfy
        the criteria for demonstrating novel AV perception
        attacks while simultaneously showing the real impact on
        the overall AV safety? Likewise, is it possible to
        find a provably secure defense against sensor attacks
        that do cause dangerous control outcomes?}
\end{tcolorbox}

\section{Compute Environment}
\label{sec:pipeline_3}

The compute environment plays a vital role in AVs to guarantee the real-time completion and accuracy of the perception, planning, and control tasks~\cite{liu2020computing}. Generally, it maintains the integrity of the communication middleware, real-time scheduling, and CPU/GPU resources.

\subsection{AV Software Stack}
Autoware is an ``end-to-end" open-source software solution for autonomous driving~\cite{autoware-foundation}. It offers a comprehensive suite of tools and libraries for perception, planning, and control~\cite{autoware-RA, kato2018autoware}. A similar tool is Baidu Apollo, which is also open-sourced and aiming at level 4/5 autonomy~\cite{apollo}. However, incorrect implementations and misuse of APIs, concurrency, and memory expose many software vulnerabilities~\cite{garcia2020comprehensive}, which can leak privacy information and impact the AV's safety. One example is a cache side-channel attack can be used to infer the location of an AV that runs the adaptive Monte-Carlo localization (AMCL) algorithm~\cite{luo2020stealthy}.

The AVGuardian system is a testament to the importance of software defenses. It addresses overprivilege issues in AV software, a scenario where a software module has more permissions than it requires, making it a potential weak point for attackers. By identifying and rectifying such instances, AVGuardian enforces stricter permission controls, thereby narrowing potential attack avenues \cite{hong2020avguardian}. Firmware code injection is another type of attack for the compute environment which could impact the perception and control modules of the AV~\cite{hallyburton2023securing}. By modifying the sensor data in real-time, the attacker achieves emergency braking or collision outcomes when running LiDAR-based perception and camera-LiDAR fusion perception in Apollo 7.0~\cite{hallyburton2023securing}. Besides, by simply buffering and replaying scene data, safety-critical accidents are achieved.



\subsection{Communication Middleware}
While there exist several options for AV communication middleware, ROS is by far the most commonly used platform for autonomous vehicles~\cite{quigley2009ros,macenski2022robot}. It faces significant security challenges, especially in its node and message communication systems. Within the ROS framework, there is no authentication mechanism for message passing or new node creation, which leaves it vulnerable to cyber-attacks. Attackers can use IP addresses and ports from the master node to create a new ROS node or hijack an existing one without additional authentication~\cite{jeong2017study}. This vulnerability can lead to excessive consumption of system resources, such as memory and CPU usage, thereby negatively impacting the performance of legitimate ROS nodes and potentially causing the entire autonomous driving system to collapse. Additionally, ROS message security is compromised due to the ease of intercepting and replaying messages. Attackers can monitor and record ROS message traffic through the master node's IP address and port, storing this data in a ROS bag file. This allows them to replay historical messages, disrupting ongoing communications~\cite{jeong2017study}. Furthermore, since ROS communication is based on socket communication, attackers can remotely monitor and intercept ROS messages without directly compromising the master node~\cite{lera2016cybersecurity,mcclean2013preliminary}. ROS2 is developed using the Data Distribution Service (DDS) for security support~\cite{deng2022security}, called Secure ROS2 (SROS2). However, the adversary can break the access control policies through out-of-date permission files or services. The reason is that when the access control policies are updated, SROS2 simply replaces the old permission files with the new ones, or sets up a new directory to store the new files and changes the corresponding load pointers~\cite{deng2022security}.  

\subsection{Limitations \& Takeaway}

\subsubsection{Lacking Comprehensive Code Check}

The AV software stack is complex, encompassing a vast array of functionalities from sensor fusion and machine learning algorithms to decision-making and control systems. The large volume and intricacy of the code, coupled with the dynamic and interdependent nature of the systems, make it extremely difficult to thoroughly identify and rectify all potential bugs and errors. As a result, it leaves room for vulnerabilities that could be exploited by malicious attacks, thereby posing significant risks to the safety and reliability of AVs~\cite{garcia2020comprehensive}. This scenario underscores a critical gap in the development and maintenance of AV software, highlighting the need for more robust and sophisticated methods in code analysis and security assurance. Formal verification could help~\cite{fisher2017hacms}. Extensive testing and verification is needed for the algorithm and software stack.


\subsubsection{Other Potential Vulnerabilities System Perspective}

AVs confront an array of potential vulnerabilities from a system perspective, particularly concerning timing correctness and energy consumption, which are often ignored by the emphasis on accuracy and performance. Timing correctness is crucial in AVs as even slight deviations in processing times can lead to failures in decision-making, impacting the vehicle's ability to respond to real-time road situations \cite{cho2023dynamix}. Concurrently, energy consumption poses another challenge; while efforts to minimize power usage are essential for the longevity and efficiency of AVs, they must not compromise the system's operational integrity. Techniques like Dynamic Voltage and Frequency Scaling (DVFS), used for reducing energy consumption, can, if not carefully managed, impact the predictability and consistency of system performance.


\begin{tcolorbox}[width=\columnwidth, colback=black!10, arc=3mm]
    \underline{\textsc{Research Question}$-$\ding{184}}:
        \textit{The complexity of AV software can lead to overlooked vulnerabilities, posing safety risks. Additionally, the emphasis on accuracy often neglects timing and energy efficiency aspects, which are critical for long-term safe AV performance. Can attacks on other attributes such as timing and energy lead to safety-critical failures?}
\end{tcolorbox}

\section{Algorithm}
\label{sec:pipeline_4}


On top of the compute environment, algorithms are working with various sensing data to find the optimal navigation decision. To achieve this, the car should first track the trajectory of surrounding vehicles and predict/plan the future trajectory of the ego vehicle. Given that an erroneous navigation decision can directly lead to a fatal accident, we need to ensure safety from possible attacks or malfunctions arising from the driving environment.

As deep learning technology advances, recent AV algorithms have evolved to use Deep Neural Networks (DNNs). Accordingly, there is significant research at the algorithm level that demonstrates a strong adversary exploiting the probabilistic nature of DNNs to intentionally cause safety-critical failures of AV decision-making processes \cite{yang2021robust}. 
 
\subsection{Tracking}
3D Multi-object tracking (MOT) is achieved by capturing object motion across detection results in consecutive frames~\cite{wu20213d}. Due to its ability to interpret historical context, tracking is increasingly used as a defensive strategy against attacks targeting AV perception systems. For example, PercepGuard~\cite{man2023person} checks the consistency of tracked objects with their respective category characteristics, acting as a countermeasure against misclassification attacks. In a similar vein, 3D-TC2~\cite{you2021temporal} utilizes tracking and prediction of objects' future positions to check temporal consistency with detected objects, effectively countering LiDAR spoofing attacks.

Concurrently, advanced attacks specifically designed for tracking algorithms have emerged. Jia \emph{et al.}~\cite{jia2020robust} introduced an adversarial attack and defense algorithm that manipulates the location and size of bounding boxes to boost digital domain robustness. Likewise, Muller \emph{et al.}~\cite{muller2022physical} have devised an innovative tracker hijacking method. This technique modifies the trajectory of a target object by either shifting it onto an unrelated path or completely altering its intended trajectory. A key feature of this method is the creation of physically plausible attacks, avoiding unrealistic manipulations, such as moving objects into the sky.

Moreover, tracking data is highly susceptible to malfunctions due to challenges like occlusion, dynamic and unpredictable object movements, and variations in the ego vehicle’s perspective. The absence of depth information in vision data often leads to misinterpretations. As a result, continuous research into attacks, defenses, and potential malfunctions is essential for ensuring the safety and security of AV systems.

\subsection{Navigational Safety}
In the planning module, AV systems are centered around finding the most efficient routes to their destinations. This involves merging data from detection and tracking systems to formulate future trajectories and actions. In this realm, Reinforcement Learning (RL) is notably effective, optimizing behavior across varied driving conditions while focusing on maximizing predefined rewards. Consequently, RL-based planning algorithms have increasingly become targets for adversaries. For instance, He \emph{et al.}~\cite{he2023toward} strategically manipulate observed states and environmental dynamics to mislead RL-based decision-making systems.

Taking a different direction, ACERO~\cite{song2023discovering} unveils a novel attack strategy wherein an adversary orchestrates their vehicle movements to endanger a targeted vehicle. This technique involves crafting an optimal adversarial driving maneuvers by considering safety requirements and physical constraints for a successful attack execution.

Considering that navigation decisions are more closely tied to a vehicle's active decision-making than any other module, a heightened level of safety awareness is essential. Mo \emph{et al.}~\cite{mo2022mcts_planning}, for instance, showcase the integration of risk assessment and safe policy search in RL agents. In a similar context, Li \emph{et al.}~\cite{li2023pomdp_planning} contribute to a more realistic risk assessment in multi-AV environments by taking into account the various driving intentions of surrounding vehicles, such as cut-ins and overtakings.

\subsection{Robustness to Domain Shifts}
One critical problem in self-driving vehicles is ensuring consistent object detection or segmentation performance across different domain shifts such as weather. Ensuring the safety of self-driving vehicles against adverse weather conditions such as rain, fog, or snow is a difficult yet important problem to avoid unexpected behavior in such conditions. The importance of this problem has seen an emergence of datasets that include different weather conditions such as BDD100K~\cite{yu2020bdd100k}, Cityscapes~\cite{cordts2015cityscapes} and its rainy~\cite{hu2019depth} and foggy~\cite{sakaridis2018semantic} variants, KITTI~\cite{geiger2012we} and its rainy~\cite{mirza2021robustness} and foggy~\cite{mirza2021robustness} variants, and NuScenes~\cite{caesar2020nuscenes}. With such datasets, some AV specific approaches have been proposed to handle different weather conditions~\cite{ahmed2021dfr,li2023domain,wang2020tent}. However, it is worth looking at advances in the more general computer vision domain to understand how such approaches may be improved.

As a case study, unsupervised domain adaptation in object detection and segmentation~\cite{oza2023unsupervised,ganin2016domain,chen2018domain,saito2019strong,hsu2020every,pan2020unsupervised,pan2022ml,sindagi2020prior,roychowdhury2019automatic,khodabandeh2019robust,zhao2020collaborative,hsu2020progressive,li2023domain} is one key area of interest due to its ability to leverage unlabeled images or synthetically generated images. These approaches include adversarial learning to learn domain invariant features~\cite{ganin2016domain,chen2018domain,saito2019strong,pan2020unsupervised,hsu2020every,sindagi2020prior}, self-training on pseudo labels~\cite{roychowdhury2019automatic,khodabandeh2019robust,zhao2020collaborative}, image-to-image translation~\cite{hsu2020progressive,ahmed2021dfr}, and learning with motion or temporal-based priors~\cite{pan2023moda,guan2021domain,xing2022domain,li2021unsupervised}, to name a few. Many of these works use simulated or real AV datasets and focus on the machine learning aspects with evaluations aimed around mIoU for segmentation and mAP for detection. These approaches from the vision domain offer much potential inspiration towards building AV perception systems that can one day be trusted to handle the whole range of realistic driving environments.

\subsection{Limitations \& Takeaway}
\subsubsection{Error Propagation in Sequential Execution of Multi-DNN}
The perception algorithm plays a pivotal role as it provides initial results to subsequent modules. Given that state-of-the-art perception is not entirely accurate in its predictions, there is a risk of error propagation to further ``downstream" modules. To address this, some algorithms incorporate strategies to validate perception results. These strategies involve statistically calculating the propagated error of bounding boxes \cite{Weng2022_Affinipred} or running algorithms at the raw sensor data level \cite{cho2023adopt}. However, accurately computing data uncertainty, intra-model uncertainty, and inter-model uncertainty remains a complex challenge. This complexity necessitates a more comprehensive understanding to ensure a robust risk assessment.

\subsubsection{Limited Coverage on Complex and Diverse Real-World Driving Environment}
Self-driving companies train foundational models for different tasks and deploy these models in cars via post-optimization \cite{wayve, scale}. A significant challenge arises from the limited datasets used in training these models, which may not encompass the full spectrum of real-world road environments. This limitation often results in a distribution shift in the inputs fed into AV systems, leading to erroneous outputs.

To mitigate these shift issues, methods such as fine-tuning, domain generalization, and test-time adaptation are being explored \cite{ahmed2021dfr,li2023domain,wang2020tent}. These methods aim to lessen the impact of domain shifts and ensure that systems remain up-to-date. However, since machine learning models must be optimized before deployment, this optimization process inevitably limits model size, consequently affecting their learning capabilities. Therefore, a crucial decision lies in balancing the trade-off between optimization and learning capability, and determining which model (deployed vs. foundational) should be adjusted, and at what point in time.

\subsubsection{Lack of AV Specific Solutions for Domain Shift Invariance}
To guarantee the safety of AVs, the perception algorithm must be robust to distributional shifts such as poor weather or other environmental conditions that a vehicle may run into~\cite{ahmed2021dfr,li2023domain,wang2020tent,chen2018domain,pan2020unsupervised,pan2022ml,hsu2020every,sindagi2020prior,roychowdhury2019automatic,khodabandeh2019robust,zhao2020collaborative,hsu2020progressive}. The robustness against different distribution-shifts is a defense against a ``natural" attacker, where the threat vector is to identify the worst realistic driving conditions for the perception algorithm. Such work includes papers that are marketed more overtly towards AVs~\cite{ahmed2021dfr,li2023domain,wang2020tent} and others that are more general, domain-adaptation focused but evaluated on AV-related datasets~\cite{chen2018domain,pan2020unsupervised,pan2023moda,hsu2020every,sindagi2020prior,roychowdhury2019automatic,khodabandeh2019robust,zhao2020collaborative,hsu2020progressive,guan2021domain,xing2022domain,li2021unsupervised}.
We highlight two interesting trends that could be useful to consider for further research.

The first is that these approaches generally focus on the RGB image domain and use general techniques that do not make strong assumptions about the AV's setting. For example, incorporating general state-of-the-art vision techniques such as domain adaptation with multi-sensor fusion networks (assuming of the presence of LiDAR) may lead to more robust feature extraction specifically for AVs. 
Traffic signs also have less variation than ImageNet~\cite{russakovsky2015imagenet} classes (e.g., traffic signs are limited to rigid shapes and are usually never upside-down whereas a dog has a lot more variation in its shape). This may lead to AV specific solutions to extracting robust, domain invariant features, for example by extending approaches in hierarchical classification where the root classifier determines the shape and the leaf classifiers determine the specific class~\cite{chandrasekaran2019rearchitecting}. Such an approach could bias a model towards features that are more invariant to weather conditions and may work better on traffic sign classification than more general datasets like ImageNet~\cite{russakovsky2015imagenet}.

Second, these works evaluate mostly on the basis of mAP or mIoU, which are more vision-aligned metrics and do not necessarily capture the safety risk of the overall system that may be using these ML models. The real-world ramifications of the robustness (or lack thereof) may require AV-specific metrics that take into account the full system and estimates the end-to-end safety implications of any misaligned predictions. For example, incorrectly detected objects could be weighted by their distance to the AV, with closer objects being penalized harsher since farther objects can still be avoided if detected properly in a near-future frame. Specific misclassifications can also be less severe than others (e.g., mistaking a stop sign for a speed limit sign vs. mistaking a cat for a dog). Finally, the performance of a perception algorithm can be measured in the context of how it performs within end-to-end AV system (i.e., planning and control).

\begin{tcolorbox}[width=\columnwidth, colback=black!10, arc=3mm]
    \underline{\textsc{Research Question}$-$\ding{185}}:
        \textit{How can we reinforce robustness and reliability, both intra-module and inter-module, within our pipeline to ensure preparedness for increasingly complex and unpredictable driving scenarios? Moreover, can such preparation increase defenses against a strong adversary?}
\end{tcolorbox}

\section{Environment Interaction}
\label{sec:pipeline_5}


\subsection{Legal Considerations}
In the context of an AV attack impacting human safety, the legal implications are multifaceted and complex. When such an attack results in a collision or similar incident, determining liability involves analyzing several factors, including the nature of the ML system, the intent behind the adversarial input, and existing legal frameworks.

First, the nature of the ML system plays a crucial role. For AVs, the manufacturer could be held liable under product liability laws if the system is found to be defective or not reasonably safe. Vladeck argues that traditional product liability principles can extend to AI systems, particularly when they fail to perform as safely as an ordinary consumer would expect~\cite{vladeck2014machines}. Similarly, Calo suggests that manufacturers might be held strictly liable for damages caused by their products, regardless of fault or negligence~\cite{calo2015robotics}.

Second, the source and intent of the adversarial input are crucial. If the AV attack is traced back to a specific entity or individual, they could be held liable under tort law for intentionally causing harm. As Kerr points out, if a person knowingly manipulates data to cause an AI system to malfunction, they might be liable for any resulting damages or injuries under theories of intentional torts or cyber-trespass~\cite{kerr2016norms}.

Finally, the existing legal frameworks and their adaptation to emerging technologies are paramount in such scenarios. As Pagallo discusses, the evolution of AI challenges traditional legal concepts of liability and responsibility, necessitating a reevaluation of legal frameworks to accommodate these technological advances~\cite{pagallo2013laws}. This view is also supported by Solaiman, who highlights the need for new legal paradigms to address the unique challenges posed by ML systems~\cite{solaiman2017legal}, and Widen and Koopman, who propose that the AV industry should cooperate with government regulators to slowly introduce policy that would improve public comfort with AVs~\cite{widen2022autonomous}.

\subsection{Regulatory Standards}
There are several standards that govern the safety and security
of autonomous vehicles. The most notable standard is ANSI/UL 4600,
which outlines a standard for autonomous product safety~\cite{UL4600}.
While it does not define pass/fail criteria for safety, it does
provide guidance for how to ensure safety is applied to the design,
testing, validation, human-machine interaction, and other processes.
Alongside this standard, the NHTSA requires reporting of crashes
that involve vehicles that are equipped with autonomous
systems~\cite{nhtsa_standing_order}. Moreover, there are
standards from ISO that define the functional safety of vehicles
in the event of system failures~\cite{ISO26262}, as well as
the intended functionality of vehicle systems in the absence
of faults~\cite{ISO21448}. The IEEE also provides a standard
set of scenarios that AVs should be able to safely handle~\cite{IEEE2846}.

For security, the primary standard for security of road vehicles
is ISO/SAE 21434~\cite{ISO21434}. There are other standards
that concern the standard for vehicle cybersecurity management
systems~\cite{UNR155}, and auditing the cybersecurity of
vehicles~\cite{ISO5112}. Outside of vehicles, there are
additional standards for security, such as ISO27000~\cite{ISO27000}
and ANSI/ISA 62443~\cite{ANSI62443}.

\subsection{Limitations \& Takeaway}

\subsubsection{Laws Must Balance Safety Without Slowing Tech Growth}
Due to recent safety concerns regarding AVs, there is a
growing distrust for their public use. In response to these
concerns, governments (local or national) may begin to impose
strict sanctions on their operation. While this response is
well-intentioned and likely required in order to regain
faith in the public deployment of AVs, there is room for
the legislation that follows to strongly discourage the
development of the tech. If it is too strict, then the barriers
it enforces may deincentivize further investment and ultimately
kill the technology on the stop. Future legislation must
carefully consider the intended outcomes once enacted,
and support should be provided to AV companies to ensure
they can continue development.

\subsubsection{Ad Hoc Provisions of AV ``Driving Licenses"}
The existing standards primarily serve the purpose of
L2/L3 AV features. Presently, when a L4 AV hits the road,
local governments provide a license to the autonomous driver
based on {\em ad hoc} standards that adapt as issues arise.
This method of providing fixes as issues arise is against
the goals of safety engineering, which seeks to eliminate risk
before damages occur. Presently, issuing licenses to L4
AVs require stricter regulation and a stronger safety culture.

\subsubsection{Modeling Human Behavior is Hard}
Despite the legal necessity to demonstrate functional safety,
especially when human-life is at risk, it is extremely difficult
to model the behavior of other people. This includes how other
vehicles may respond to the ego vehicle's control, how aggressive
cyclists may behave on the road, or how pedestrians may act
during a road crossing scenario. No matter the situation,
people can sometimes do seemingly random things and catch
the AV off-guard. As a result, despite any best effort, it is
not currently practical to completely guarantee functional safety
requirements.

\begin{tcolorbox}[width=\columnwidth, colback=black!10, arc=3mm]
    \underline{\textsc{Research Question}$-$\ding{186}}:
        \textit{Future legislation will need to consider
        how and when AV software can be granted a license
        to autonomously operate in specific environments.
        How can this be provided to ensure the public does
        not involuntarily partake in a dangerous experiment,
        while simultaneously encourage the further development
        of AV technology?}
\end{tcolorbox}

\section{End-to-End Evaluation Examples}
\label{sec:pipeline_6}
Before moving on to our future recommendations, we discuss
some cases where there is a substantial attempt at
end-to-end testing of AVs. One of the earliest
models of AVs is from Mobileye's RSS formal model for
providing safety assurances~\cite{shalev2017formal}. RSS provides
a model based on physics for assuring the safety of 
the vehicle. While the model is abstract and requires
exact precision of the measurements of the vehicle's
surroundings to ensure the safety
of the vehicle, its a step in the right direction.
Other work investigate impacts on safety due to failures
in the perception stack to fill this gap~\cite{fremont2020formal}.
There are also additions to end-to-end testing that
attempt to test for rare events~\cite{okelly2018scalable}, using adaptive
stress testing to pursue stronger coverage of
scenarios~\cite{corso2019adaptive}, and using fuzzing to discover
bugs in the AV stack~\cite{kim2022drivefuzz}.

The major gap in end-to-end validation testing of AV
safety is in the simulation tools. There is a substantial
gap between AVs in simulation compared to real-world
AVs, otherwise known as the sim-to-real gap. Overcoming
this drawback is still an area of active inquiry. Current
solutions primarily seek to cross the sim-to-real gap via
transferability of training in simulation to real-world
environments~\cite{salvato2021crossing,daza2023sim,trentsios2022overcoming}.
These usages of sim-to-real do not seek to validate the system safety.

\section{Future Research Recommendations}
\label{sec:future}

{\bf Recommendation \#1}: For future research in AV security, a key recommendation is to focus on comprehensive end-to-end testing strategies that encompass the entire system's operation, from perception to control, to ensure robustness and safety in diverse real-world scenarios. This would include the future creation of universal metrics that may be used across all problems to accurately weigh the strengths and weaknesses of various approaches. These metrics may measure various aspects of the pipeline in order to demonstrate where attacks/defenses are most effective, as well as the end result so that one may see the threat that an attack has on functional safety.

{\bf Recommendation \#2}: 
Effectively utilizing the data from surrounding vehicles and integrating it is essential to gain a more comprehensive understanding of the driving environment. Cooperative perception and planning strategies can significantly enhance safety and security, particularly in scenarios where a vehicle is compromised or its sensors are disrupted. This approach ensures more reliable and accurate sensing outcomes, even in challenging circumstances. Moreover, it's crucial to consider that drivers themselves can inadvertently be a source of safety and security issues. Even without malicious intent, their unavoidable actions may resemble attack-like behavior to other vehicles in specific traffic situations. Therefore, in addition to the AV technology stacks illustrated in \autoref{fig:overview}, monitoring and understanding driver behaviors can play a pivotal role in preventing AVs from exacerbating potential hazards.

{\bf Recommendation \#3}:
The licenses of AVs should have more concrete and strict regulation for when and how they are provided and enforced. This would include clearer standards for the sort of scenarios that the AVs can provide provable functional safety for. Moreover, these AVs should be required to have well-regulated privacy policies with respect to the data they collect on the road. Packaging privacy requirements with the safety requirements of an AV license would provide an encouraging solution to both issues at once. What remains to be determined is where the regulation ``sweet spot" is for strictly enforcing an AV license. Government should seek to avoid deincentivizing the development of AVs, so finding this balance is crucial.



\bibliographystyle{ACM-Reference-Format}
\bibliography{bib}


\end{document}